\documentclass[11pt,a4paper]{article}
\usepackage[hyperref]{emnlp2018}
\usepackage{times}
\usepackage{url}
\usepackage{latexsym}
\usepackage{multirow}
\usepackage{tabularx}
\usepackage{graphicx,url,microtype,soul,needspace,paralist,booktabs,rotating}
\newcolumntype{Y}{>{\raggedright\arraybackslash}X}
\newcolumntype{R}{>{\raggedleft\arraybackslash}X}
\newcolumntype{C}{>{\centering\arraybackslash}X}
\usepackage{xcolor}
\usepackage[normalem]{ulem}
\usepackage{amsmath}
\useunder{\uline}{\ul}{}

\usepackage{url}

\aclfinalcopy 

\title{Deep learning for language understanding of mental health concepts derived from Cognitive Behavioural Therapy}

\author{Lina Rojas-Barahona$^1$, Bo-Hsiang Tseng$^1$, Yinpei Dai$^1$, Clare Mansfield$^2$ \\
\textbf{Osman Ramadan$^1$, Stefan Ultes$^1$, Michael Crawford $^3$ and} \\
\textbf{Milica~Ga{\v s}i{\' c}}$^1$\\
$^\textbf{1}\ $University of Cambridge,
$^\textbf{2}\ $CM Insight,
$^\textbf{3}\ $Imperial College London\\
\texttt{mg436@cam.ac.uk}}

\date{}

\begin{document}
\maketitle
\begin{abstract}
In recent years, we have seen deep learning and distributed representations of words and sentences make impact on a number of natural language processing tasks, such as similarity, entailment and sentiment analysis. Here we introduce a new task: understanding of mental health concepts derived from Cognitive Behavioural Therapy (CBT). We define a mental health ontology based on the CBT principles, annotate a large corpus where this phenomena is exhibited and perform understanding using deep learning and distributed representations. Our results show that the performance of deep learning models combined with word embeddings or sentence embeddings significantly outperform non-deep-learning models in this difficult task. This understanding module will be an  essential component of a statistical dialogue system delivering therapy.

\end{abstract}

\section{Introduction}\label{sec:intro}


Promotion of mental well-being is at the core of the action plan on mental health 2013--2020 of the World Health Organisation~(WHO)~\cite{who13} and of the European Pact on Mental Health and Well-being of the European Union~\cite{eu08}. The biggest potential breakthrough in fighting mental illness would lie in finding tools for early detection and preventive intervention~\cite{insc06}. The WHO action plan stresses the importance of health policies and programmes that not only meet the need of people affected by mental disorders but also protect mental well-being. The emphasis is on early evidence-based non-pharmacological intervention, avoiding institutionalisation and medicalisation.  What is particularly important for successful intervention is the frequency with  which the therapy can be accessed~\cite{half02}.  This gives automated systems a huge advantage over conventional therapies, as they can be used continuously with marginal extra cost. Health assistants that can deliver therapy, have gained great interest in recent years~\cite{bickmore2005establishing,fitzpatrick2017delivering}. These systems however are largely based on hand-crafted rules. On the other hand, the main research effort in statistical approaches to conversational systems has focused on limited-domain information seeking dialogues~\cite{swsy06,gepi11,gayo14,fash16,lmrj16,wikz17}.

In this paper we introduce a new task: understanding of mental health concepts derived from Cognitive Behavioural Therapy~(CBT). We present an ontology that is formulated according to Cognitive Behavioural Therapy principles. We label a high quality mental health corpus, which exhibits targeted psychological phenomena.
We use the whole unlabelled dataset to train distributed representations of words and sentences. We then investigate two approaches for classifying the user input according to the defined ontology.
The first model involves a convolutional neural network (CNN) operating over distributed words representations. The second involves a gated recurrent network (GRU) operating over distributed representation of sentences.
Our models perform significantly better than chance and for instances with a large number of data they reach the inter-annotator agreement.  This understanding module will be an  essential component of a statistical dialogue system delivering therapy.

 
The paper is organised as follows. In Section~\ref{sec:back} we give a brief background of the statistical approach to dialogue modelling, focusing on dialogue ontology and natural language understanding. In Section~\ref{sec:rel} we review related work in the area of automated mental-health assistants.  The sections that follow represent the main contribution of this work: a CBT ontology in Section~\ref{sec:ontology}, a labelled dataset in Section~\ref{sec:data}, and models for language understanding in Section~\ref{sec:model}. We present the results in Section~\ref{sec:results} and our conclusion in Section~\ref{sec:conclusion}.

\section{Background}\label{sec:back}

A dialogue system can be treated as a trainable statistical model suitable for goal-oriented information seeking dialogues \cite{youn02a}. In these dialogues, the user has a clear goal that he or she is trying to achieve and this involves extracting particular information from a back-end database.  A structured representation of the database, the \emph{ontology} is a central element of a dialogue system.  It defines the concepts that the dialogue system can understand and talk about. Another critical component is the natural language understanding unit, which takes textual user input and detects presence of the ontology concepts in the text.



\subsection{Dialogue ontology}

Statistical approaches to dialogue modelling have been applied to relatively simple domains. These systems interface databases of up to $1000$ entities where each entity has up to $20$ properties, i.e. \emph{slots}~\cite{cuay09}.
There has been a significant amount of work in spoken language understanding focused on exploiting large knowledge graphs in order to improve coverage~\cite{tjwh12,hhtu13}. Despite these efforts, little work has been done on mental health ontologies for supporting cognitive behavioural therapy on dialogue systems. Available medical ontologies follow a symptom-treatment categorisation and are not suitable for dialogue or natural language understanding~\cite{ontopsy2017,Hofmann2014576,WANG201834}.



\subsection{Natural language understanding}

Within a dialogue system, a natural language understanding unit extracts meaning from user sentences. Both classification \cite{mgjk09} and sequence-to-sequence \cite{ypzy14,mdyb15} models have been applied to address this task. 




Deep learning architectures that exploit distributed word-vector representations have been successfully applied to different tasks in natural language understanding, such as semantic role labelling, semantic parsing, spoken language understanding, sentiment analysis or dialogue belief tracking~\cite{collobert2011natural,kim2014convolutional,kalchbrenner2014convolutional,lemi14,barahona2016exploiting,mrksicNBT}. 


In this work we consider understanding of mental health concepts of as a classification task.  To facilitate this process, we use distributed representations. 


\section{Related work}\label{sec:rel}

The aim of building an automated therapist has been around since the first time researchers attempted to build a dialogue system \cite{weiz66}.  Automated health advice systems built to date typically rely on expert coded rules and have limited conversational capabilities \cite{ijmiRojas-BarahonaG09,vrbs12,rbtb13,ricc14,dabd14,ribp16}. One particular system that we would like to highlight is an affectively aware virtual therapist \cite{ribp16}. This system is based on Cognitive Behavioural Therapy and the system behaviour is scripted using VoiceXML. There is no language understanding: the agent simply asks questions and the user selects answers from a given list. The agent is however able to interpret hand gestures, posture shifts, and facial expressions. Another notable system \cite{dabd14} has a multi-modal perception unit which captures and analyses user behaviour for both behavioural understanding and interaction. The measurements contribute to the indicator analysis of affect, gesture, emotion and engagement.  Again, no statistical language understanding takes place and the behaviour of the system is scripted. The system does not provide therapy to the user but is rather a tool that can support healthcare decisions (by human healthcare professionals).  

The Stanford Woebot chat-bot proposed by \cite{fitzpatrick2017delivering} is designed for delivering CBT to young adults with depression and anxiety. It has been shown that the interaction with this chat-bot can significantly reduce the symptoms of depression when compared to a group of people directed to a read a CBT manual. The conversational agent appears to be  effective in engaging the users. However, the understanding component of Woebot has not been fully described. The dialogue decisions are based on decision trees.  At each node, the user is expected to choose one of several predefined responses. Limited language understanding was introduced at specific points in the tree to determine routing to subsequent conversational nodes. Still, one of the main deficiencies reported by the trial participants in \cite{fitzpatrick2017delivering} was the inability to converse naturally. 
Here we address this problem by performing statistical natural language understanding.


\section{CBT ontology}\label{sec:ontology}

To define the ontology we draw from principles of Cognitive Behavioural Therapy (CBT).  This is one of the best studied psychotherapeutic interventions, and the most widely used psychological treatment for mental disorders in Britain \cite{napt13}.  There is evidence that CBT is more effective than other forms of psychotherapy \cite{toli10}. Unlike other, longer-term, forms of therapy such as psychoanalysis, CBT can have a positive effect on the client within a few sessions. Also, due to it being highly structured, it is more easily amenable by computer interpretation.  This is why we adopted CBT as the basis of our work. 

Cognitive Behavioural Therapy is derived from Cognitive Therapy model theory \cite{beck76,brse79}, which postulates that our emotions and behaviour are influenced by the way we think and by how we make sense of the world.  The idea is that, if the client changes the way he or she thinks about their problem, this will in turn change the way he or she feels, and behaves.

A major underlying principle of CBT is the idea of cognitive distortions, and the value in challenging them.  In CBT, clients are helped to test their assumptions and views of the world in order to check if they fit with reality. When clients learn that their perceptions and interpretations are distorted or unhelpful they then work on correcting them.  Within the realm of cognitive distortion, CBT identifies a number of specific self-defeating thought processes, or thinking errors. There is a core of around 10 to 15 thinking errors, with their exact titles having some fluidity.  A strong component of CBT is teaching clients to be able to recognize and identify the thinking errors themselves, and ultimately discard the negative thought processes and `re-think' their problems.  

We consider the main analytical step in this therapy:  an adequate decoding of these `thinking error' concepts, and the identification of the key emotion(s) and the situational context of a particular problem.  Therefore, our ontology consists of \emph{thinking errors}, \emph{emotions}, and \emph{situations}. 

\subsection{Thinking errors}
Notwithstanding slight variations in number and terminology, the list of \emph{thinking errors} is fairly well standardised in the CBT literature.  
We present one such list in Table~\ref{tab:thinkingerr}.  
However, it is important to note that there is a fair degree of overlap between different \emph{thinking errors}, for example, between \emph{Jumping to Negative Conclusions} and \emph{Fortune Telling}, or between \emph{Disqualifying the Positives} and \emph{Mental Filtering}.  In addition, within the data used -- and as is likely to be the case in any data of spontaneous expressions of psychological upset -- a single problem can exhibit several \emph{thinking errors} simultaneously.  Thus, the situation is much more challenging than in simple information-seeking dialogues, where ontologies are typically clearly defined and there is no or very little overlap between concepts.

\begin{table*}[ht!]
\centering
\scalebox{0.82}{
\begin{tabular}{lll}
\hline
Thinking Error &Frequency& Exhibited by... \\ \hline
Black and white (or all or nothing) thinking &$20.82\%$& \begin{tabular}[c]{@{}l@{}}Only seeing things in absolutes\\
No shades of grey\end{tabular} \\ \hline
Blaming &$8.05\%$& \begin{tabular}[c]{@{}l@{}}Holding others responsible for your pain \\ Not seeking to understand your own responsibility in situation\end{tabular} \\ \hline
Catastrophising &$11.87\%$& \begin{tabular}[c]{@{}l@{}}	Magnifying a (sometimes minor) negative event \\into potential disaster
\end{tabular} \\ \hline
Comparing &$3.27\%$& \begin{tabular}[c]{@{}l@{}}	Making dissatisfied comparison of self versus others
\end{tabular} \\ \hline
Disqualifying the positive &$6.15\%$& \begin{tabular}[c]{@{}l@{}}Dismissing/discounting positive aspects \\of a situation or experience
\end{tabular} \\ \hline
Emotional reasoning &$13.31\%$& \begin{tabular}[c]{@{}l@{}}Assuming feelings represent fact.
\end{tabular} \\ \hline
Fortune telling &$25.70\%$& Predicting how things will be, unduly negatively \\ \hline
Jumping to negative conclusions &$44.16\%$& \begin{tabular}[c]{@{}l@{}}Anticipating something will turn out badly, \\with little evidence to support it\end{tabular} \\ \hline
Labelling &$10.51\%$& \begin{tabular}[c]{@{}l@{}}Using negative, sometimes highly coloured, language \\to describe self or other\\
Ignoring complexity of people\end{tabular} \\ \hline
\begin{tabular}[c]{@{}l@{}}

Low frustration tolerance\\ "I can't bear it"\end{tabular} &$16.03\%$& \begin{tabular}[c]{@{}l@{}}Assuming something is intolerable,\\ rather than difficult to tolerate or a temporary discomfort\end{tabular} \\ \hline
\begin{tabular}[c]{@{}l@{}}

Inflexibility\\ 			 "should/need/ought"\end{tabular} &$8.08\%$& \begin{tabular}[c]{@{}l@{}}Having rigid beliefs \\ about how things  or people ‘must’ or ‘ought to’ be\end{tabular} \\ \hline
Mental filtering &$5.50\%$& \begin{tabular}[c]{@{}l@{}}Focusing on the negative\\ Filtering out all positive aspects of a situation\end{tabular} \\ \hline
Mind-reading &$14.60\%$& \begin{tabular}[c]{@{}l@{}} 	Assuming others think negative things \\ or have negative motives and intentions\end{tabular}
 \\ \hline
Over-generalising &$12.69\%$& \begin{tabular}[c]{@{}l@{}}Generalising negatively,\\ using words like ‘always’, ‘nobody’, ‘never’, etc\end{tabular} \\ \hline
Personalising &$5.85\%$& \begin{tabular}[c]{@{}l@{}}Interpreting events as being related to you personally and \\ overlooking other factors \end{tabular} \\ \hline
\end{tabular}%
}
\caption{Taxonomy for \emph{thinking errors} and how they are exhibited.}
\label{tab:thinkingerr}
\end{table*}

\subsection{Emotions}
In addition to \emph{thinking errors}, we define a set of \emph{emotions}.  We mainly focus on negative emotions, relevant to people in psychological distress. In CBT, emotions tend to be divided into positive and negative, or helpful\slash healthy and unhelpful\slash unhealthy emotions \cite{brwi10}. 
The set of emotions for this work evolved over time in the early days of annotation.  Although we initally agreed to focus on `unhealthy' emotions, as defined by CBT, there seemed also to be a place for the `healthy' emotion \emph{Grief/sadness}.  Overall, the list of emotions used was drawn from a number of sources, including CBT literature, the annotators' own knowledge of what they work with in psychological therapy, and the common emotions that were seen emerging from the data early on in the process. Note that more than one emotion might be expressed within an individual problem -- for example \emph{Depression} and \emph{Loneliness}. The list of \emph{emotions} is given in Table~\ref{tab:emo}.  

\begin{table*}[ht!]
\begin{minipage}{0.6\textwidth}
\scalebox{0.82}{
\begin{tabular}{lll}
\hline
Emotion & Frequency & Exhibited by ... \\ \hline
Anger (/frustration) &$14.76\%$ & \begin{tabular}[c]{@{}l@{}}	Feelings of frustration, annoyance,\\irritation,  resentment, fury, outrage
\end{tabular} \\ \hline
Anxiety &$63.12\%$& \begin{tabular}[c]{@{}l@{}}Any expression of fear, worry or anxiety\end{tabular} \\ \hline
Depression &$20.72\%$& \begin{tabular}[c]{@{}l@{}}Feeling down, hopeless, joyless, negative \\ about self and/or life in general\end{tabular} \\ \hline
Grief/sadness &$5.70\%$& \begin{tabular}[c]{@{}l@{}}Feeling sad, upset, bereft \\ in relation to a major loss\end{tabular} \\ \hline
Guilt &$3.37\%$& \begin{tabular}[c]{@{}l@{}} Feeling blameworthy \\ for a wrongdoing or something not done
\end{tabular} \\ \hline
Hurt &$19.88\%$& \begin{tabular}[c]{@{}l@{}}Feeling wounded and/or badly treated\end{tabular} \\ \hline
Jealousy &$3.12\%$& \begin{tabular}[c]{@{}l@{}}Antagonistic feeling towards another \\ either wish to be like or to have what they have\end{tabular} \\ \hline
Loneliness &$7.41\%$ & \begin{tabular}[c]{@{}l@{}}Feeling of alone-ness, isolation, friendlessness,\\ not understood by anyone\end{tabular} \\ \hline
Shame &$5.68\%$ & \begin{tabular}[c]{@{}l@{}}Feeling distress, humiliation, disgrace \\in relation to own behaviour or feelings\end{tabular} \\ \hline
\end{tabular}}
\caption{Taxonomy for \emph{emotions} and how they are exhibited.}
\label{tab:emo}
\end{minipage}
\begin{minipage}{0.5\textwidth}
\begin{center}
\scalebox{0.82}{
\begin{tabular}{cl}\hline
Situation&Frequency\\\hline
Bereavement &$2.65\%$\\  
Existential & $21.93\%$ \\
Health & $10.61\%$\\
Relationships & $67.58\%$\\
School/College &$8.28\%$\\
Work &$6.10\%$\\
Other & $5.53\%$\\\hline
\end{tabular}}
\caption{Taxonomy for \emph{situations}.}
\label{tab:situ}
\end{center}
\end{minipage}
\end{table*}

\subsection{Situations}
While our main emphasis was on \emph{thinking errors} and \emph{emotions}, we also defined a  small set of \emph{situations}.  The list of \emph{situations} again evolved during the early days of annotation, with a longer original list being reduced down, for simplicity. Again, it is possible for more than one situation (for example \emph{Work} and \emph{Relationships}) to apply to a single problem. The considered \emph{situations} are given in Table~\ref{tab:situ}.

\begin{figure}
\centering
\includegraphics[width=80mm,trim={0 0 0 0},clip]{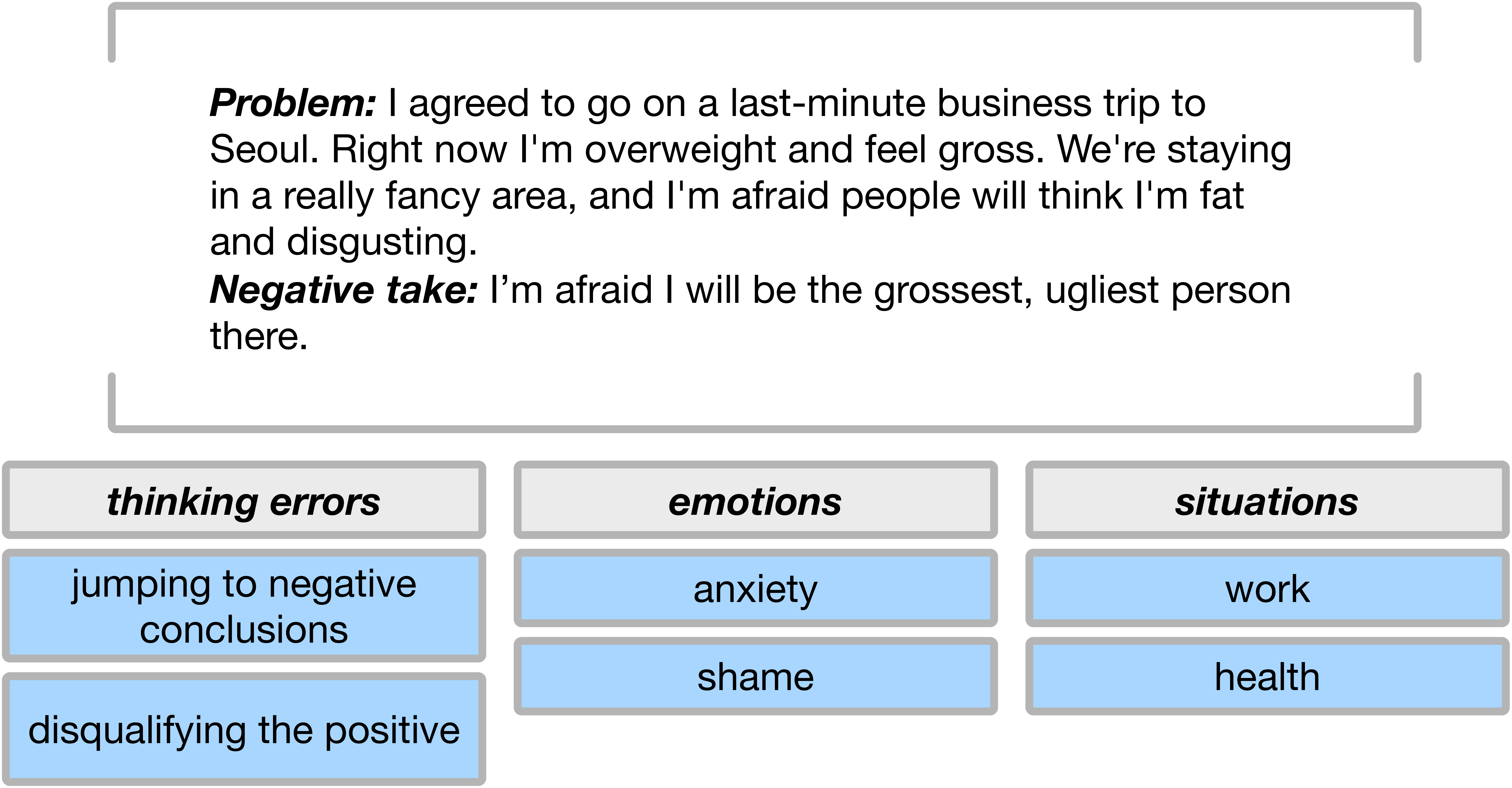}
\caption{An example of an annotated Koko post.}
\label{fig:koko}
\vspace{-0.5cm}
\end{figure}

\section{The corpus}\label{sec:data}

The corpus consists of $500$K written posts that users anonymously posted on the Koko platform\footnote{\url{https://itskoko.com/}}. This platform is based on the peer-to-peer therapy proposed by \cite{mosp15}. In this set-up, a user anonymously posts their problem (referred to as the \textit{problem}) and is prompted to consider their most negative take on the problem (referred to as the \textit{negative take}). Subsequently, peers post responses that attempt to offer a re-think and give a more positive angle on the problem.  When first developed, this peer-to-peer framework was shown to be more efficacious than expressive writing, an intervention that is known to improve physical and emotional well-being \cite{mosp15}. Since then, the app developed by Koko has collected a very large number of posts and associated responses. Initially, any first-time Koko user would be given a short introductory tutorial in the art of `re-thinking'\slash `re-framing' problems (based on CBT principles), before being able to use the platform.  This however changed over time, as the age of the users decreased, and a different tutorial, emphasizing empathy and optimism, was used (less CBT-based than the `re-thinking'). Most of the data annotated in this study was drawn from the earlier phase.  
Figure~\ref{fig:koko} gives an annotated post example.

\subsection{Annotation}
A subset of posts was annotated by two psychological therapists using a web annotation tool that we developed.  The annotation tool allowed annotators to have a quick view of the posts, showing up to 50 posts per page, to navigate through posts, to check pending posts and to annotate them by adding or removing \emph{thinking errors}, \emph{emotions} and \emph{situations}. All annotations were stored in a MySQL database.

\begin{figure}[htbp]
\centering
\includegraphics[width=80mm]{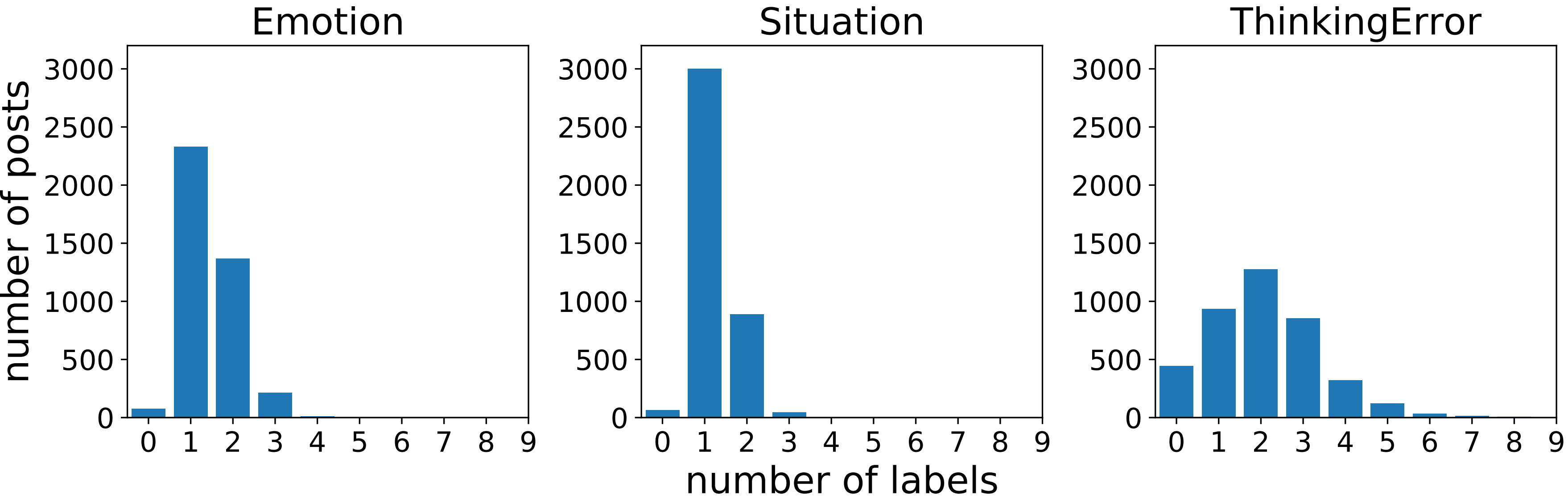}
\caption{Distribution of posts for each category.}
\label{fig:post_dist}
\vspace{-0.5cm}
\end{figure}

Initially $1000$ posts were analysed. These were used to define the ontology. Then $4035$ posts were labelled with \emph{thinking errors}, \emph{emotions} and \emph{situations}. It takes an experienced psychological therapist about one minute to annotate one post. Note that the same post can exhibit multiple \emph{thinking errors}, \emph{emotions} and \emph{situations}, which makes the whole process more complex. We randomly selected 50 posts and calculated the inter-annotator agreement. The inter-annotator agreement was calculated using a contingency table for thinking error, emotion and situation, showing agreement and disagreement between the two annotators. Then, Cohen's kappa was calculated discounting the possibility that the agreement may happen by chance. The result is shown in Table~\ref{tab:agreement}.  The main reason for the low agreement in \emph{thinking errors} (61\%) is due to the unbounded number of \emph{thinking errors} per post. In other words, the annotators typically have three or four \emph{thinking errors} in common but one of them might have detected one or two more. Still, the agreement is much higher than chance, so we think that while challenging, it is possible to build a classifier for this task.
The distributions of labelled posts with multiple sub-categories for three super-categories are shown in Figure \ref{fig:post_dist}

\begin{table}
\begin{center}
\begin{tabular}{c|ccc}
\hline
\scriptsize{Concept} & \scriptsize{Thinking error} & \scriptsize{Situation}&\scriptsize{Emotion}\\
\hline
\scriptsize{Kappa} & \scriptsize{$0.61 \pm 0.09$} & \scriptsize{$0.92 \pm 0.08$} & \scriptsize{$0.90 \pm 0.07$}\\
\hline
\end{tabular}
\caption{Cohen's kappa with a $95\%$ confidence interval}
\label{tab:agreement}
\vspace{-0.7cm}
\end{center}
\end{table}

\section{Deep learning model}\label{sec:model}

\subsection{Distributed representations}
The task of decoding \emph{thinking errors} and \emph{emotions} is closely related to the task of sentiment analysis. In sentiment analysis we are concerned with positive or negative sentiment expressed in a sentence. Detecting thinking errors or emotions could be perceived as detecting different kinds of negative sentiment.  Distributed representations of words, sentences and documents have gained success in sentiment detection and similarity tasks \cite{lemi14,maas2011learning,kiros2015skip}. 
A key advantage of these representations is that they can be obtained in an unsupervised manner, thus allowing exploitation of large amounts of unlabelled data. This is precisely what we have in our set-up, where only a small portion of our posts is labelled.

We utilise GloVe \cite{pennington2014glove} word vectors, which have previously achieved competitive results in a similarity task. We train the word vectors on the whole dataset and then use a convolutional neural network (CNN) to extract features from posts where words are represented as vectors.


We also consider distributed representation of sentences. A particularly competitive model is the skip-thought model, which is obtained from an encoder-decoder model that tries to reconstruct the surrounding sentences of an encoded passage \cite{kiros2015skip}. On similarity tasks it outperfoms the simpler doc2vec model \cite{lemi14}. An approach that represents vectors by weighted averages of word vectors and then modifies them using PCA and SVD outperforms skip-thought vectors \cite{arlt17}. This method however does not do well on a sentiment analysis task due to down-weighting of words like ``not''. As these often appear in our corpus, we chose skip-thought vectors for investigation here.

\begin{figure}[ht!]
\centering
\includegraphics[width=75mm]{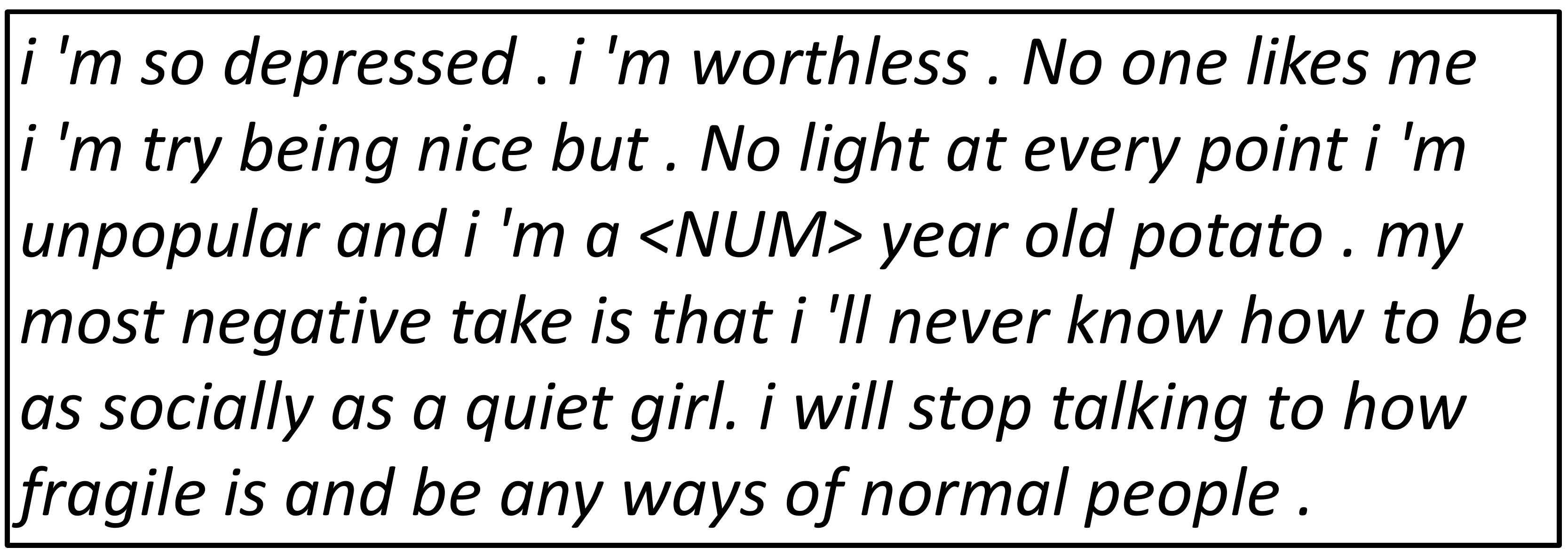}
\caption{An example of a generated post using skip-thought vectors initialised with "I'm so depressed".}
\vspace{-0.5cm}
\label{fig:Post_generation}
\end{figure}

The skip-thought model allows a dense representation of the utterance. We train skip-thought vectors using the method described in~\cite{kiros2015skip}.  The automatically generated post shown in Fig~\ref{fig:Post_generation} demonstrates that skip-thought vectors can convey the sentiment well in accordance to context. We then train a gated recurrent unit (GRU) network using the skip-thoughts as input. 

\subsection{Convolutional neural network model}
\begin{figure}[ht!]
\vspace{-0.3cm}
\centering
\includegraphics[width=75mm]{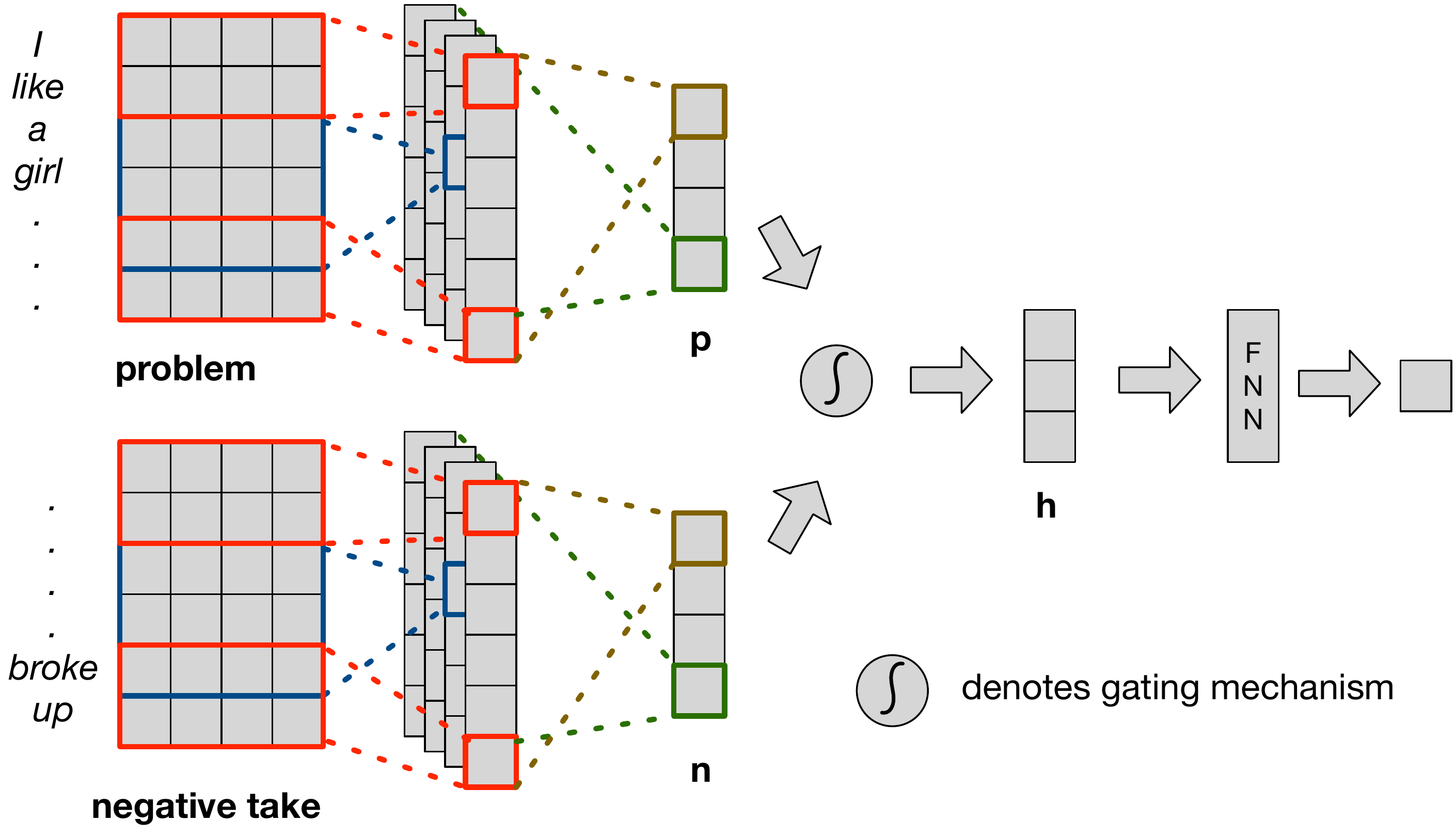}
\caption{CNN with gating mechanism.}
\label{fig:CNN}
\vspace{-0.5cm}
\end{figure}

The convolutional neural network (CNN) model is preferred over a recurrent neural network (RNN) model, because the posts are generally too long for an RNN to maintain memory over words.
The convolutional neural network (CNN) used in this work is inspired by \cite{kim2014convolutional} and operates over pre-trained GloVe embeddings of dimensionality~$d$.  As shown in Fig~\ref{fig:CNN}, the network has two inputs, one for the \textit{problem} and the other for the \textit{negative take}. These are represented as two tensors. A convolutional operation involves a filter $\mathbf{w} \in \mathcal{R}^{ld}$ which is applied to $l$ words to produce the feature map. Then, a max-pooling operation is applied to produce two vectors: $\mathbf{p}$ for \textit{problem} and $\mathbf{n}$ for \textit{negative take}. The reason for this is that the \emph{negative take} is usually a summary of the post, carrying stronger sentiment~(see Figure~\ref{fig:koko}). We use a gating mechanism to combine $\mathbf{p}$ and $\mathbf{n}$ as follows:
\setlength\abovedisplayskip{1pt}
\setlength\belowdisplayskip{0.5pt}
\begin{align}
  \mathbf{g}&=\sigma ( \mathbf{W}_{p}  \mathbf{p} + \mathbf{W}_{n}  \mathbf{n} + \mathbf{b} ) \\
   \mathbf{h} &= \mathbf{g} \odot \mathbf{p} + (\mathbf{1}-\mathbf{g}) \odot \mathbf{n}
\end{align} 
Here, $\sigma$ is the sigmoid function, $\mathbf{W}_p$, $\mathbf{W}_{n}$ and $\mathbf{W}$ are weight matrices, $\mathbf{b}$ is a bias term, $\mathbf{1}$ is a vector of ones, $\odot$ is the element-wise product, and $\mathbf{g}$ is the output of the gating mechanism. The extracted feature $\mathbf{h}$ is then processed with a one-layer fully-connected neural network (FNN) to perform binary classification. The model is illustrated in Fig~\ref{fig:CNN}.

\subsection{Gated recurrent unit model}
We use the gated recurrent unit~(GRU) model to process skip-thought sentence vectors, for two reasons. First, most posts contain less than 5 sentences, so a recurrent neural network is more suitable than a convolutional neural network. Second, since our corpus only comprises very limited labelled data, a GRU should perform better than a long short-term memory~(LSTM) network as it has less parameters.

Denote each post as $P=\{s_1,s_2,...,s_t,...\}$, where $s_t$ is the $t^{\text{th}}$ sentence in post~$P$. First, we use an already trained GRU to extract skip-thought embeddings $\mathbf{e_t}$ from the sentences~$s_t$. Then, taking the sequence of sentence vectors $\{\mathbf{e_1},\mathbf{e_2},...,\mathbf{e_t},...\}$ as input, another GRU is used as follows:
\setlength\abovedisplayskip{2pt}
\setlength\belowdisplayskip{2pt}
\begin{align}
\mathbf{z_t}&=\sigma ( \mathbf{W}_{z} \mathbf{h}_{t-1} + \mathbf{U}_{z}  \mathbf{e_t} + \mathbf{b_z}) \\
\mathbf{r_t}&=\sigma ( \mathbf{W}_{r} \mathbf{h}_{t-1} + \mathbf{U}_{r}  \mathbf{e_t} + \mathbf{b_r}) \\
\mathbf{\tilde{h}_t}&= \tanh ( \mathbf{W}(\mathbf{r_t} \odot \mathbf{h}_{t-1} )  + \mathbf{U}  \mathbf{e_t} + \mathbf{b_h}) \\
\mathbf{h_t} &= \mathbf{z_t} \odot \mathbf{h}_{t-1} + (\mathbf{1}-\mathbf{z_t}) \odot \mathbf{\tilde{h}_t}
\end{align}
$\mathbf{W}_z, \mathbf{U}_z, \mathbf{W}_r, \mathbf{U}_r ,\mathbf{W}, \mathbf{U}$ are recurrent weight matrices,  $\mathbf{b_z, b_r,b_h}$ are bias terms, $\odot$ is the element-wise dot product, and $\sigma$ is the sigmoid function.

\begin{figure}[htbp]
\vspace{-0.3cm}
\centering
\includegraphics[width=75mm]{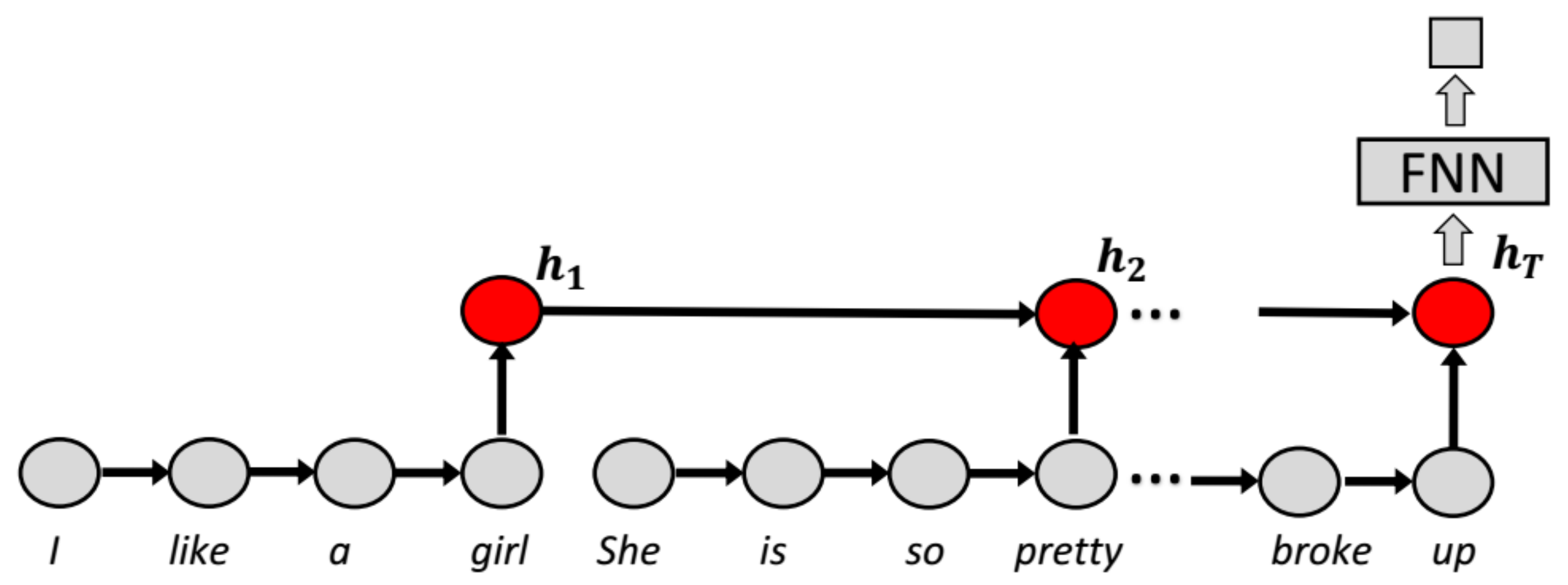}
\caption{GRU with skip-thought vectors.}
\label{fig:GRU}
\vspace{-0.5cm}
\end{figure}

Finally, the last hidden state $\mathbf{h}_{T}$ is fed into a FNN with one hidden layer of the same size as input. The model is illustrated in Fig~\ref{fig:GRU}.

\subsection{Training set-up}
\label{set-up}
We first train $100$ and $300$ dimensions for both GloVe embeddings and skip-thought embeddings using the same mechanism as in \cite{pennington2014glove,kiros2015skip}. In some posts the length of sentences is very large, so we bound the length at $5$0 words. We do not treat the \emph{problem} separately from the \emph{negative take} as the GRU will anyway put more importance on the information that comes last. We split the labelled data in a $8:1:1$ ratio for training, validation and testing in a $10$-fold cross validation for both GRU and CNN training. A distinct network is trained for each concept, i.\,e.\ one for \emph{thinking errors}, one for \emph{emotions} and one for \emph{situations}. The hidden size of the FNN is $150$.

To tackle the data bias problem, we utilise oversampling. Different ratios (1:1, 1:3, 1:5, 1:7) of positive and negative samples are explored.

We used filter windows of $2$, $3$, and $4$ with $50$ feature maps for the CNN model. For the GRU model, the hidden size is set at $150$, so that both models have comparable number of parameters. Mini-batches of size $24$ are used and gradients are clipped with maximum norm~$5$. We initialise the learning rate as $0.001$ with a decay rate of $0.986$ every 10  steps. The non-recurrent weights with a truncated normal distribution $(0, 0.01)$, and the recurrent weights with orthogonal initialisation \cite{saxe2013exact}. To overcome over-fitting, we employ dropout with rate $0.8$ and $l2$-normalisation. Both models were trained with Adam algorithm and implemented in Tensorflow \cite{girija2016tensorflow}.

\section{Results}\label{sec:results}
\subsection{Baselines}
For rule-based models, we chose a chance classifier and a majority classifier, where all the posts are treated as positive examples for each class. In addition, we trained two non-deep-learning models, the logistic regression (LR) model and the Support Vector Machine (SVM). Both of them take the bag-of-words feature as input and implemented in sklearn \cite{scikit11learn}. For completeness, we also trained 100 and 300 dimensions PV-DM document embeddings \cite{le2014distributed} as the distributed representations of the posts using the \emph{gensim} toolkit \cite{rehurek_lrec}, and employ FNNs to do the classification, the hidden size is set as 800 to ensure parameters of all deep learning models comparable. All the baseline models are trained with the same set-up as described in section \ref{set-up}.

\subsection{Analysis}
Table \ref{tab:all_models} gives the average F1 scores and the average F1 scores weighted with the frequency of CBT labels for all models under the oversampling ratio 1:1. It shows that GloVe word vectors with CNN achieves the best performance both in 100 and 300 dimensions.
\begin{table}[htbp]
\centering
\scalebox{0.7}{
\begin{tabular}{c|c|c}
\hline
Model & AVG. F1 & Weighted AVG F1 \\
\hline
Chance & 0.203$\pm$0.008   & 0.337$\pm$0.008 \\
Majority & 0.24$\pm$0.000 & 0.432$\pm$0.000   \\
LR-BOW & 0.330$\pm$0.011 & 0.479$\pm$0.008  \\
SVM-BOW &  0.403$\pm$0.000 & 0.536$\pm$0.000\\
\hline
FNN-DocVec-100d &   0.339$\pm$0.006 & 0.502$\pm$0.005\\
FNN-DocVec-300d &  0.349$\pm$0.007 & 0.508$\pm$0.005 \\
GRU-SkipThought-100d &  0.401$\pm$0.005 & 0.558$\pm$0.004\\
GRU-SkipThought-300d &  0.423$\pm$0.005 & 0.570$\pm$0.004 \\
CNN-GloVe-100d &  $\mathbf{0.443\pm0.007}$ & 0.576$\pm$0.005 \\
CNN-GloVe-300d & $0.442\pm0.007$ & $\mathbf{0.578\pm0.006}$\\
\hline
\end{tabular}}
\caption{F1 scores for all models with 1:1 oversampling}
\label{tab:all_models}
\vspace{-0.3cm}
\end{table}

\begin{table*}[ht!]
\centering
\scalebox{0.7}{
\begin{tabular}{lllcccc}
\hline
&{Freq.}             & \multirow{2}{*}{SVM-BOW} & \multicolumn{2}{c}{100d}                             & \multicolumn{2}{c}{300d}        \\
           & {Num.}&                       & \multicolumn{1}{c}{CNN-Glove} & \multicolumn{1}{c}{GRU-Skip-thought} & \multicolumn{1}{c}{CNN-Glove} & \multicolumn{1}{c}{GRU-Skip-thought} \\ \hline
\textbf{Emotion}            &                                              &                            &                          &                            &     \\
Anxiety & 2547 & 0.798$\pm$0.000 & 0.805$\pm$0.003 & 0.805$\pm$0.002 & 0.805$\pm$0.006 & $\mathbf{0.816}\pm\mathbf{0.002}$ \\
Depression & 836 & 0.564$\pm$0.000 & 0.605$\pm$0.003 & 0.568$\pm$0.001 & $\mathbf{0.611}\pm\mathbf{0.008}$ & 0.578$\pm$0.005  \\
Hurt & 802 & 0.448$\pm$0.000 & 0.505$\pm$0.007 & 0.483$\pm$0.003 & $\mathbf{0.506}\pm\mathbf{0.005}$ & 0.496$\pm$0.006  \\
Anger & 595 & 0.375$\pm$0.001 & 0.389$\pm$0.009 & 0.384$\pm$0.007 & 0.383$\pm$0.004 & $\mathbf{0.425}\pm\mathbf{0.007}$ \\
Loneliness & 299 & 0.558$\pm$0.000 & 0.495$\pm$0.008 & 0.445$\pm$0.007 & $\mathbf{0.549}\pm\mathbf{0.009}$ & 0.457$\pm$0.005  \\
Grief & 230 & 0.433$\pm$0.005 & 0.462$\pm$0.010 & 0.373$\pm$0.008 & $\mathbf{0.462}\pm\mathbf{0.008}$ & 0.382$\pm$0.005  \\
Shame & 229 & 0.220$\pm$0.000 & $\mathbf{0.304}\pm\mathbf{0.011}$ & 0.243$\pm$0.004 & 0.277$\pm$0.007 & 0.254$\pm$0.004 \\
Jealousy & 126 & 0.217$\pm$0.000 & $\mathbf{0.228}\pm\mathbf{0.012}$ & 0.159$\pm$0.004 & 0.216$\pm$0.005 & 0.216$\pm$0.009 \\
Guilt & 136 & 0.252$\pm$0.000 & $\mathbf{0.295}\pm\mathbf{0.012}$ & 0.186$\pm$0.007 & 0.279$\pm$0.014 & 0.225$\pm$0.008 \\
\hline
AVG. F1 score for \textbf{Emotion} &  & 0.429$\pm$0.001 & $\mathbf{0.454}\pm\mathbf{0.008}$ & 0.405$\pm$0.005  & $\mathbf{0.454}\pm\mathbf{0.007}$ & 0.428$\pm$0.006 \\
\hline

\textbf{Situation}            &                                              &                            &                          &                            &     \\
Relationships & 2727 & 0.861$\pm$0.000 & 0.871$\pm$0.003 & 0.886$\pm$0.001 & 0.878$\pm$0.006 & $\mathbf{0.889}\pm\mathbf{0.003}$ \\
Existential & 885 & 0.556$\pm$0.000 & 0.591$\pm$0.002 & $\mathbf{0.600}\pm\mathbf{0.005}$ & 0.594$\pm$0.007 & 0.599$\pm$0.006 \\
Health & 428 & 0.476$\pm$0.000 & $\mathbf{0.589}\pm\mathbf{0.003}$ & 0.555$\pm$0.005 & 0.585$\pm$0.008 & 0.587$\pm$0.006 \\
School\_College & 334 & 0.633$\pm$0.000 & 0.670$\pm$0.004 & 0.641$\pm$0.003 & 0.673$\pm$0.009 & $\mathbf{0.680}\pm\mathbf{0.002}$ \\
Other & 223 & 0.196$\pm$0.001 & 0.255$\pm$0.011 & 0.241$\pm$0.008 & 0.256$\pm$0.005 & $\mathbf{0.281}\pm\mathbf{0.006}$ \\
Work & 246 & 0.651$\pm$0.000 & $\mathbf{0.663}\pm\mathbf{0.004}$ & 0.572$\pm$0.006 & 0.661$\pm$0.011 & 0.639$\pm$0.006 \\
Bereavement & 107 & 0.602$\pm$0.000 & 0.637$\pm$0.021 & 0.402$\pm$0.024 & $\mathbf{0.639}\pm\mathbf{0.021}$ & 0.493$\pm$0.011  \\
\hline
AVG. F1 score for \textbf{Situation} &  & 0.568$\pm$0.000 & 0.611$\pm$0.007 & 0.557$\pm$0.007 & $\mathbf{0.612}\pm\mathbf{0.010}$  & 0.595$\pm$0.006 \\
\hline
\textbf{Thinking Error}            &                                              &                            &                          &                            &     \\
Jumping\_to\_negative\_conclusions & 1782 & 0.590$\pm$0.000 & 0.696$\pm$0.004 & 0.685$\pm$0.004 & $\mathbf{0.703}\pm\mathbf{0.005}$ & 0.687$\pm$0.002  \\
Fortune\_telling & 1037 & 0.458$\pm$0.000 & $\mathbf{0.595}\pm\mathbf{0.002}$ & 0.558$\pm$0.004 & 0.585$\pm$0.006 & 0.564$\pm$0.005 \\
Black\_and\_white & 840 & 0.395$\pm$0.000 & 0.431$\pm$0.002 & 0.437$\pm$0.004 & 0.432$\pm$0.003 & $\mathbf{0.441}\pm\mathbf{0.003}$ \\
Low\_frustration\_tolerance & 647 & 0.318$\pm$0.000 & 0.322$\pm$0.007 & 0.330$\pm$0.003 & 0.313$\pm$0.005 & $\mathbf{0.336}\pm\mathbf{0.001}$ \\
Catastrophising & 479 & 0.352$\pm$0.000 & $\mathbf{0.375}\pm\mathbf{0.002}$ & 0.358$\pm$0.005 & 0.371$\pm$0.004 & 0.364$\pm$0.003 \\
Mind-reading & 589 & 0.360$\pm$0.000 & 0.404$\pm$0.005 & 0.353$\pm$0.011 & $\mathbf{0.419}\pm\mathbf{0.006}$ & 0.356$\pm$0.007  \\
Labelling & 424 & 0.399$\pm$0.001 & 0.453$\pm$0.007 & 0.335$\pm$0.004 & $\mathbf{0.462}\pm\mathbf{0.004}$ & 0.373$\pm$0.002  \\
Emotional\_reasoning & 537 & 0.290$\pm$0.000 & $\mathbf{0.319}\pm\mathbf{0.007}$ & 0.285$\pm$0.005 & 0.306$\pm$0.006 & 0.293$\pm$0.008 \\
Over-generalising & 512 & 0.405$\pm$0.001 & 0.405$\pm$0.006 & 0.375$\pm$0.004 & $\mathbf{0.418}\pm\mathbf{0.008}$ & 0.389$\pm$0.004  \\
Inflexibility & 326 & 0.202$\pm$0.001 & 0.203$\pm$0.014 & 0.188$\pm$0.007 & $\mathbf{0.218}\pm\mathbf{0.003}$ & 0.208$\pm$0.005  \\
Blaming & 325 & 0.209$\pm$0.001 & $\mathbf{0.304}\pm\mathbf{0.007}$ & 0.264$\pm$0.002 & 0.277$\pm$0.003 & 0.274$\pm$0.004 \\
Disqualifying\_the\_positive & 248 & 0.146$\pm$0.000 & 0.194$\pm$0.007 & 0.176$\pm$0.005 & 0.187$\pm$0.003 & $\mathbf{0.195}\pm\mathbf{0.005}$ \\
Mental\_filtering & 222 & 0.088$\pm$0.000 & 0.142$\pm$0.007 & 0.150$\pm$0.001 & 0.141$\pm$0.002 & $\mathbf{0.155}\pm\mathbf{0.003}$ \\
Personalising & 236 & 0.212$\pm$0.000 & 0.230$\pm$0.012 & 0.220$\pm$0.005 & $\mathbf{0.236}\pm\mathbf{0.004}$ & 0.221$\pm$0.005  \\
Comparing & 132 & 0.242$\pm$0.000 & $\mathbf{0.289}\pm\mathbf{0.014}$ & 0.177$\pm$0.008 & 0.255$\pm$0.009 & 0.227$\pm$0.007 \\
\hline
AVG. F1 score for \textbf{Thinking Error} &  & 0.311$\pm$0.000 & $\mathbf{0.358}\pm\mathbf{0.007}$   & 0.326$\pm$0.005 & 0.355$\pm$0.0050 &0.339$\pm$0.004 \\

\hline
\hline
AVG. F1 score&  & 0.403$\pm$0.000 & $\mathbf{0.443}$$\pm\mathbf{0.007}$ & 0.401$\pm$0.005 & 0.442$\pm$0.007 & 0.423$\pm$0.005 \\
AVG. F1 score weighted with Freq.&  & 0.536$\pm$0.000 & 0.576$\pm$0.005 & 0.558$\pm$0.004 & $\mathbf{0.578}$$\pm\mathbf{0.006}$ & 0.570$\pm$0.004 \\
\hline
\end{tabular}}
\caption{F1 score of the models trained with embeddings with dimensionality of 300 and 100 respectively.}
\label{tab:results}
\vspace{-0.5cm}
\end{table*}

\begin{table*}[ht!]
\centering
\scalebox{0.7}{
\begin{tabular}{lcccc}
\hline
label & precision & recall & F1 score & accuracy \\
\hline
Anxiety & 0.739$\pm$0.007 & 0.884$\pm$0.005  & 0.805$\pm$0.006  & 0.729$\pm$0.012 \\
Depression & 0.538$\pm$0.010 & 0.708$\pm$0.005  & 0.611$\pm$0.008  & 0.813$\pm$0.010 \\
Hurt & 0.428$\pm$0.005 & 0.620$\pm$0.004  & 0.506$\pm$0.005  & 0.763$\pm$0.011 \\
Anger & 0.313$\pm$0.005 & 0.491$\pm$0.000  & 0.383$\pm$0.004  & 0.769$\pm$0.012 \\
Loneliness & 0.479$\pm$0.010 & 0.643$\pm$0.008  & 0.549$\pm$0.009  & 0.923$\pm$0.006 \\
Grief & 0.437$\pm$0.013 & 0.490$\pm$0.000  & 0.462$\pm$0.008  & 0.937$\pm$0.005 \\
Shame & 0.219$\pm$0.008 & 0.378$\pm$0.004  & 0.277$\pm$0.007  & 0.891$\pm$0.007 \\
Jealousy & 0.170$\pm$0.002 & 0.296$\pm$0.012  & 0.216$\pm$0.005  & 0.935$\pm$0.006 \\
Guilt & 0.221$\pm$0.014 & 0.378$\pm$0.008  & 0.279$\pm$0.014  & 0.936$\pm$0.008 \\
Relationships & 0.847$\pm$0.005 & 0.912$\pm$0.007  & 0.878$\pm$0.006  & 0.829$\pm$0.011 \\
Existential & 0.516$\pm$0.008 & 0.700$\pm$0.004  & 0.594$\pm$0.007  & 0.789$\pm$0.009 \\
Health & 0.520$\pm$0.010 & 0.668$\pm$0.005  & 0.585$\pm$0.008  & 0.900$\pm$0.006 \\
School\_College & 0.570$\pm$0.009 & 0.821$\pm$0.008  & 0.673$\pm$0.009  & 0.934$\pm$0.004 \\
Other & 0.209$\pm$0.004 & 0.331$\pm$0.007  & 0.256$\pm$0.005  & 0.894$\pm$0.007 \\
Work & 0.601$\pm$0.015 & 0.733$\pm$0.006  & 0.661$\pm$0.011  & 0.955$\pm$0.003 \\
Bereavement & 0.567$\pm$0.029 & 0.733$\pm$0.008  & 0.639$\pm$0.021  & 0.979$\pm$0.002 \\
Jumping\_to\_negative\_conclusions & 0.643$\pm$0.005 & 0.775$\pm$0.004  & 0.703$\pm$0.005  & 0.711$\pm$0.009 \\
Fortune\_telling & 0.486$\pm$0.006 & 0.737$\pm$0.004  & 0.585$\pm$0.006  & 0.733$\pm$0.010 \\
Black\_and\_white & 0.330$\pm$0.003 & 0.625$\pm$0.003  & 0.432$\pm$0.003  & 0.657$\pm$0.011 \\
Low\_frustration\_tolerance & 0.222$\pm$0.005 & 0.531$\pm$0.002  & 0.313$\pm$0.005  & 0.631$\pm$0.028 \\
Catastrophising & 0.291$\pm$0.005 & 0.509$\pm$0.000  & 0.371$\pm$0.004  & 0.796$\pm$0.012 \\
Mind-reading & 0.343$\pm$0.008 & 0.540$\pm$0.002  & 0.419$\pm$0.006  & 0.783$\pm$0.014 \\
Labelling & 0.376$\pm$0.004 & 0.597$\pm$0.003  & 0.462$\pm$0.004  & 0.853$\pm$0.007 \\
Emotional\_reasoning & 0.241$\pm$0.006 & 0.417$\pm$0.004  & 0.306$\pm$0.006  & 0.748$\pm$0.017 \\
Over-generalising & 0.337$\pm$0.009 & 0.548$\pm$0.002  & 0.418$\pm$0.008  & 0.808$\pm$0.014 \\
Inflexibility & 0.162$\pm$0.002 & 0.336$\pm$0.006  & 0.218$\pm$0.003  & 0.807$\pm$0.012 \\
Blaming & 0.218$\pm$0.002 & 0.381$\pm$0.005  & 0.277$\pm$0.003  & 0.841$\pm$0.009 \\
Disqualifying\_the\_positive & 0.125$\pm$0.002 & 0.365$\pm$0.008  & 0.187$\pm$0.003  & 0.808$\pm$0.016 \\
Mental\_filtering & 0.087$\pm$0.001 & 0.386$\pm$0.009  & 0.141$\pm$0.002  & 0.741$\pm$0.026 \\
Personalising & 0.179$\pm$0.003 & 0.345$\pm$0.007  & 0.236$\pm$0.004  & 0.871$\pm$0.009 \\
Comparing & 0.257$\pm$0.009 & 0.253$\pm$0.009  & 0.255$\pm$0.009  & 0.952$\pm$0.003 \\
\hline
\end{tabular}}
\caption{Precision, recall, F1 score and accuracy for 300 dim CNN-GloVe with oversampling ratio 1:1}
\label{tab:CNN-GloVe300}
\vspace{-0.5cm}
\end{table*}
Table~\ref{tab:results} shows the F1-measure of the compared models that detect \emph{thinking errors}, \emph{emotions} and \emph{situations} under the $1:1$ oversampling ratio. We only include the results of the best performing models, SVMs, CNNs and GRUs, due to limited space. The results show that both models outperform SVM-BOW in larger embedding dimensions. Although SVM-BOW is comparable to 100 dimensional GRU-Skip-thought in terms on average F1, in all other cases CNN-GloVe and GRU-Skip-thought overshadow SVM-BOW. We also find that CNN-GloVe on average works better than GRU-Skip-thought, which is expected as the space of words is smaller in comparison to the space of sentences so the word vectors can be more accurately trained. While the CNN operating on 100 dimensional word vectors is comparable to the CNN operating on 300 dimensional word vectors, the GRU-Skip-thought tends to be worse on 100 dimensional skip-thoughts, suggesting that sentence vectors generally need to be of a higher dimension to represent the meaning more accurately than word vectors.

\begin{figure}[ht!]
\centering
\includegraphics[width=70mm]{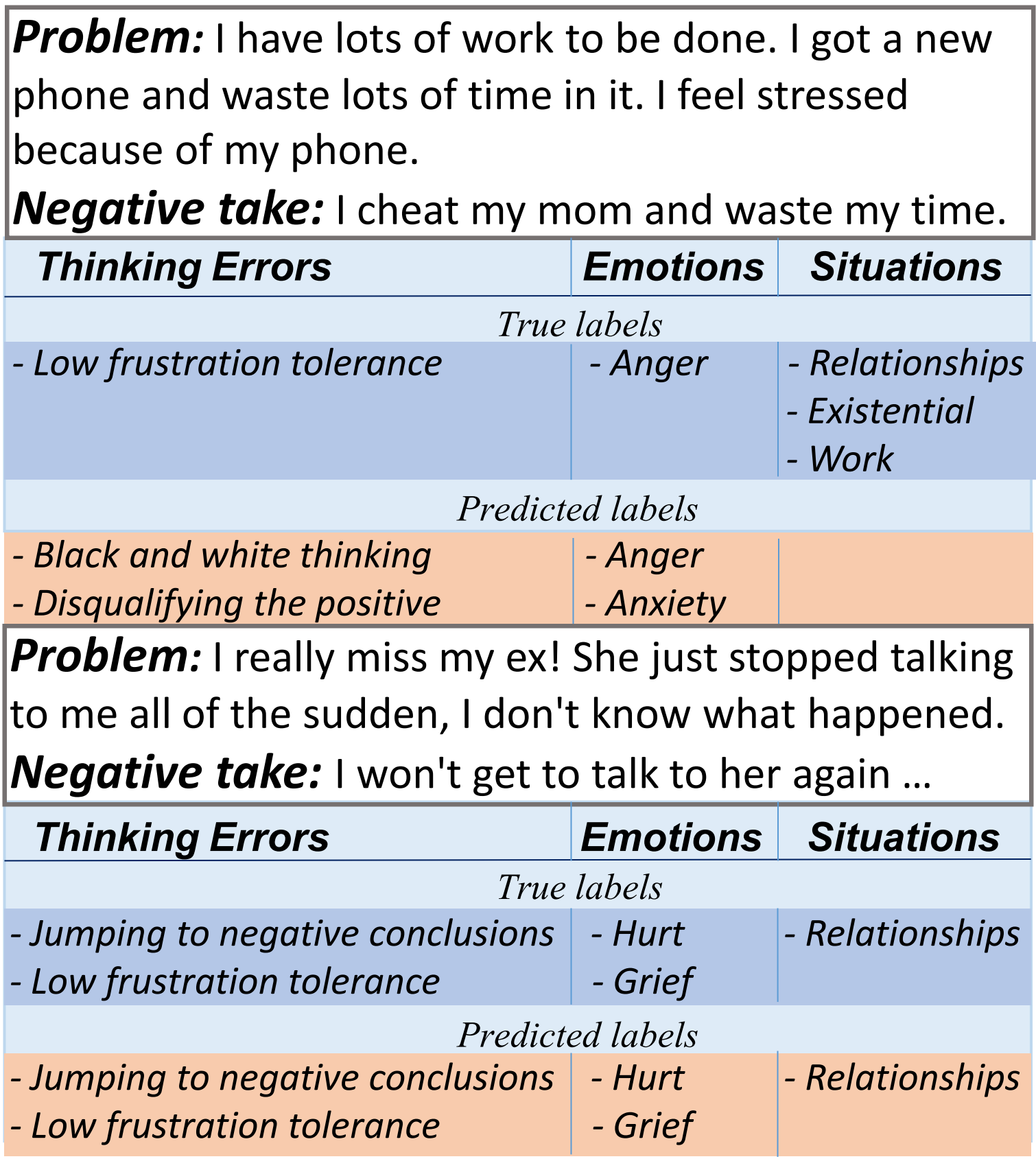}
\caption{predictions of posts by 300 dim CNN-GloVe}
\label{fig:prediction}
\vspace{-0.5cm}
\end{figure}

Table \ref{tab:CNN-GloVe300} shows a more detailed analysis of the 300 dimensional CNN-GloVe performance, where both precision and recall are presented, indicating that oversampling mechanism can help overcome the data bias problem.  To illustrate the capabilities of this model, we give samples of two posts and their predicted and true labels in Figure \ref{fig:prediction}, which shows that our model discerns the classes reasonably well even in some difficult cases. 

\begin{figure}[ht!]
\vspace{-0.0cm}
\centering
\includegraphics[width=80mm]{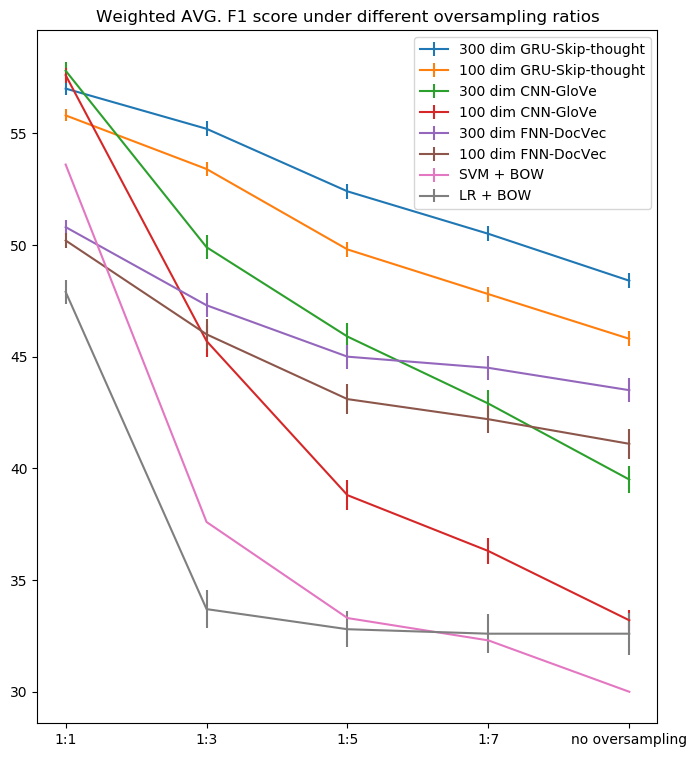}
\caption{Weighted AVG. F1 for different models}
\label{fig:oversampling}
\vspace{-0.5cm}
\end{figure}

Figure~\ref{fig:oversampling} gives the comparative performance of two models under different oversampling ratios. While oversampling is essential for both models, GRU-Skip-thought is less sensitive to lower oversampling ratios, suggesting that skip-thoughts can already capture sentiment on the sentence level. Therefore, including only a limited ratio of positive samples is sufficient to train the classifier. Instead, models using word vectors need more positive data to learn sentence sentiment features.


\section{Conclusion}\label{sec:conclusion}
We presented an ontology based on the principles of Cognitive Behavioural Therapy. We then annotated data that exhibits psychological problems and computed the inter-annotator agreement.

We found that classifying \emph{thinking errors} is a difficult task as suggested by the low inter-annotator agreement. We trained GloVe word embeddings and skip-thought embeddings on $500$K posts in an unsupervised fashion and generated distributed representations both of words and of sentences. We then used the GloVe word vectors as input to a CNN and the skip-thought sentence vectors as input to a GRU. 
The results suggest that both models significantly outperform a chance classifier for all \emph{thinking errors}, \emph{emotions} and \emph{situations} with CNN-GloVe on average achieving better results. 

Areas of future investigation include richer distributed representations, or a fusion of distributed representations from word-level, sentence-level and document-level, to acquire more powerful semantic features.  We also plan to extend the current ontology with its focus on \emph{thinking errors}, \emph{emotions} and \emph{situations} to include a much lager number of concepts.  The development of a statistical system delivering therapy will moreover require further research on other modules of a dialogue system.
\section*{Acknowledgements}
This work was funded by EPSRC project Natural speech Automated Utility for Mental health (NAUM), award reference EP/P017746/1. The authors would also like to thank anonymous reviewers for their valuable comments. The code is available at \texttt{https://github.com/YinpeiDai/NAUM}
\vspace{-0.5cm}
\bibliography{emnlp2018}
\bibliographystyle{acl_natbib_nourl}

\end{document}